\documentclass[accepted]{uai2022} 

\usepackage[american]{babel}
\usepackage{courier}
\usepackage{caption}
\usepackage{subcaption}
\usepackage{booktabs}
\usepackage{multirow}

\usepackage{natbib} 
\usepackage{mathtools} 
\usepackage{booktabs} 
\usepackage{tikz} 

\usepackage[capitalize,noabbrev]{cleveref}
\title{Selection Collider Bias in Large Language Models}

\author{\href{mailto:<emcmilin@cs.stanford.edu>?Subject=Your UAI 2022 paper}{Emily~McMilin}{}}

\affil{%
    Independent Researcher
}
\begin{document}
\maketitle
\begin{abstract}

In this paper we motivate the causal mechanisms behind sample selection induced collider bias (selection collider bias) that can cause Large Language Models (LLMs) to learn unconditional dependence between entities that are unconditionally independent in the real world. We show that selection collider bias can become amplified in underspecified learning tasks, and although difficult to overcome, we describe a method to exploit the resulting spurious correlations for determination of when a model may be uncertain about its prediction. We demonstrate an uncertainty metric that matches human uncertainty in tasks with gender pronoun underspecification on an extended version of the Winogender Schemas evaluation set, and we provide an online demo where users can apply our uncertainty metric to their own texts and models.
\end{abstract}

\section{Introduction}
This paper investigates models trained to estimate the conditional distribution: $P(Y | X , Z)$ from datasets composed of cause: $X$, effect: $Y$, and covariates: $Z$, where $Z$ is the cause of sample selection bias in the training dataset. We argue that datasets without some form of selection bias are rare, as almost all datasets are subsampled representations of a larger population, yet few are sampled with randomization.

Sample selection bias occurs when some mechanism, observed or not, causes samples to be included or excluded from the dataset. This is distinct from both confounder and collider bias. The former can occur when two variables have a common cause, and the latter can occur when two variables have a common effect. Correcting for confounding bias requires that one condition upon the common cause variable; conversely correcting for collider bias requires that one does not condition upon the common effect \cite{Pearl09}.

\begin{figure}
     \centering
     \begin{subfigure}[b]{0.31\columnwidth}
         \centering
         \includegraphics[width=\columnwidth]{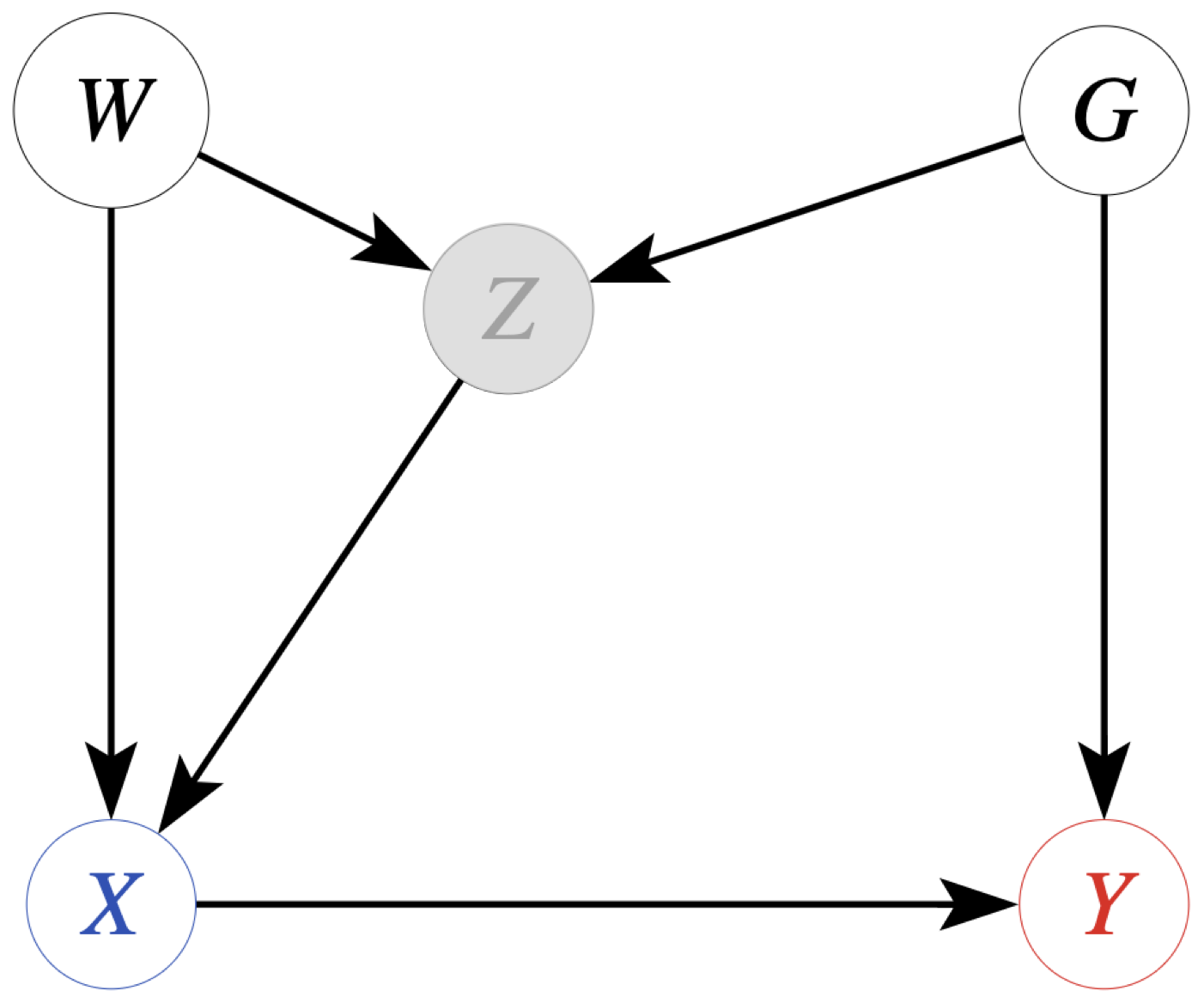}
         \caption{$G$ optionally observed and sample selection bias not taking place. } 
         \label{fig-DAG-w-G}
     \end{subfigure}
     \hfill
     \begin{subfigure}[b]{0.31\columnwidth}
         \centering
         \includegraphics[width=\columnwidth]{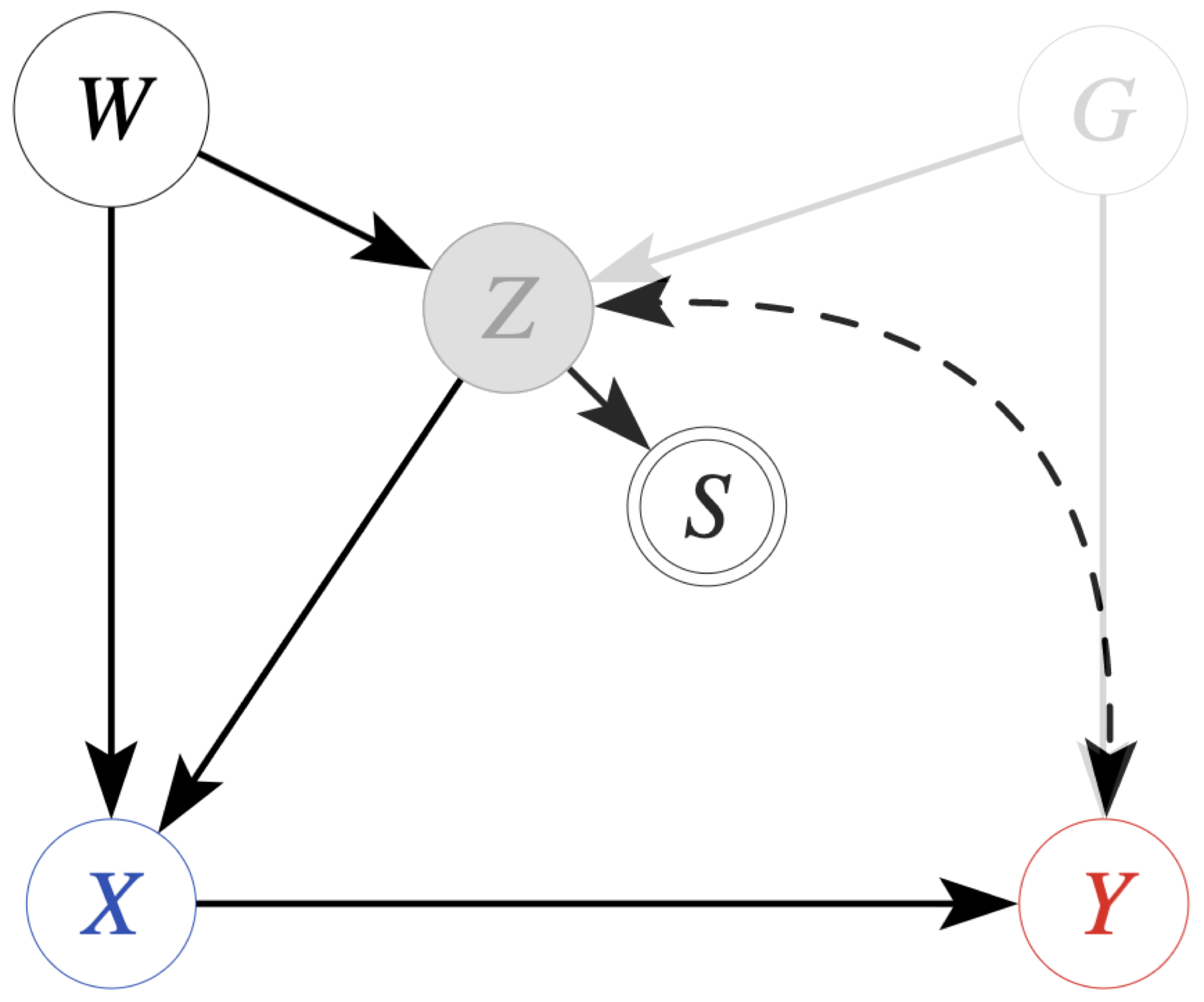}
         \caption{$G$ is unobserved, with selection bias from $S\!\!=\!\!1$ for samples in dataset.}
         \label{fig-DAG-w-S-node}
     \end{subfigure}
     \hfill
     \begin{subfigure}[b]{0.31\columnwidth}
         \centering
         \includegraphics[width=\columnwidth]{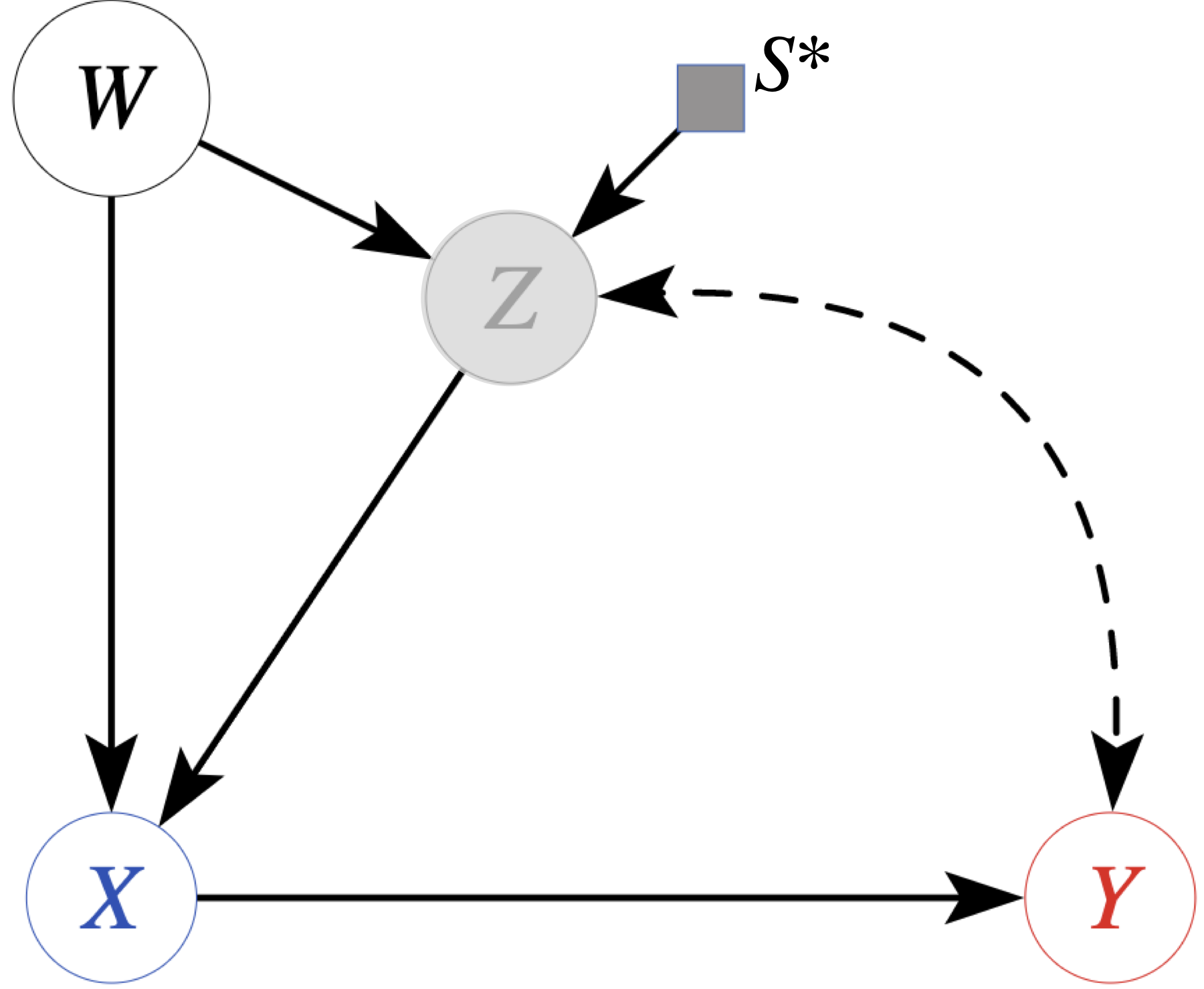}
         \caption{Causal mechanism for $Z$ varies from population $\Pi$ to $\Pi^*$. }
         \label{fig-DAG-w-trans}
     \end{subfigure}
        \caption{Proposed data generating process for a range of NLP datasets, with text-based variables: $X\!$ as gender-neutral text, $Y\!$ as a gender-identifying word (often pronoun), and symbolic variables: $W\!$, as gender-neutral entities (such as  \textit{time}  and \textit{location}), $Z$ as  \textit{access to resources}, and finally $G$ as gender. While only $X$ and $Y$ are the actual text in the dataset, both symbolic variables $W\!$ and $G$ can appear in text form in the dataset (such as the country name `Mali' and the word `man'), and $Z$ is never observed in datasets but can be partially measured with external census data.}
        \label{fig-DAGs}
        \vskip -0.1in

\end{figure}

While sample selection bias can take many forms, the type of selection bias that interests us here is that which involves more than one variable (observed or not), whose common effect results in selection bias. Such relationships can be compactly represented in causal directed acyclic graphs (DAGs), for example illustrated in \cref{fig-DAG-w-G}, which we will motivate shortly. The absence of arrows connecting variables in causal DAGs encodes assumptions, for example that $W\!$ and $G$ in \cref{fig-DAG-w-G} are stochastically independent of one another. The arrows from both $W\!$ and $G$ to $Z\!$ encodes the assumption that $Z\!$ is a common effect of $W\!$ and $G$.

In \cref{fig-DAG-w-S-node}, the twice-encircled node $S$ represents some mechanism that can cause samples to be selected for the dataset. To represent the process of dataset formation, one must condition on $S$, thus inducing its ancestor $Z$ into a collider bias relationship between $W\!$ and $G$.

We will use the term \textit{selection collider bias} to refer to circumstances such as this one, when the selection bias mechanism induces a collider bias relationship in the dataset, that would not have been there otherwise. Beyond posing a risk to out-of-domain generalizability, selection collider bias can result in models that lack even `internal validity', as the associations learned from the data represent the statistical dependences induced by the dataset formation and not the data itself \cite{Griffith2020}.

\section{Outline}\label{outline}
This work is a continuation of our prior work in \cite{emily-spurious}, where we demonstrated spurious correlations between gender pronouns and real-world gender-neutral entities like \textit{time}  and \textit{location}, on BERT \cite{BERT} and RoBERTa \cite{RoBERTa} large pre-trained models. Here we extend the work with further exploration into the causal mechanisms behind the selection bias effect, 
an investigation of methods to overcome the induced selection bias, and ultimately a demonstration of how selection collider bias induced spurious correlations can be exploited for this purpose.

\section{Masked Gender Task}


In \cite{emily-spurious}, we desired an underspecified learning task to probe spurious associations that may remain otherwise hidden in the presence of highly predictive features. We developed what we called Masked Gender Task (MGT), a special case of Masked Language Modeling (MLM) objective, that uses a heuristic to build underspecified learning tasks by masking common gender-identifying words for prediction (see \cref{gender-id-words}). 

Although we intentionally obscure gender for the MGT, we argue that it is not an implausible occurrence that during MLM pre-training, gender-identifying words are masked for prediction in otherwise gender-neutral sequences. At inference time, the prediction of gender-identifying words or labels from gender-neutral text is common to many downstream tasks such as text classification, dialog generation, machine translation from genderless to more gendered languages, or any task requiring gendered predictions from gender-underspecified features.

We grounded our experimentation in two data source types: Wikipedia-like and Reddit-like. The DAGs in \cref{fig-DAGs} represent our relevant assumptions for the data generating processes for these data sources, detailed in \cite{emily-spurious}, and briefly revisited below. 

\subsection{Example Data Generating Processes}

The objective of the MGT is to predict masked out gender-identifying words, $Y$, based on a gender-neutral input text, $X$. We assume that in MLM pre-training, the MGT objective naturally occurs, such that input sequences include words about gender-neutral entities $W\!$, such as birth\textit{place}, birth\textit{date}, or gender-neutral \textit{topics} of online forums, yet exclude $G$, gender-identifying words or concepts.  This is represented in  \cref{fig-DAG-w-S-node}, where $G$ is replaced with a doubled headed arrow to indicate that it is unobserved in the gender-neutral input text, $X$. As mentioned above, the symbolic variable $Z$ represents \textit{access to resources} that may be gender unequal. Particularly in underspecified tasks like that of the MGT, we hypothesize that $Z$ entangles the learned relationships for $W\!$ and $G$.

%
%
%

Having described the variables of our assumed data generating processes, we now describe the cause-and-effect relationships. The absence of arrows connecting variables in \cref{fig-DAGs} encodes assumptions, for example that $W\!$ and $G$ are both independent variables and causes of $Z$. This relationship is instantiated in data sources like Wikipedia written \textit{about} people as follows, $Z$ has become increasingly less gender dependent as the \textit{date} approaches more modern times, but not evenly in every \textit{place}. In data sources like Reddit written \textit{by} people, the $W \rightarrow Z \leftarrow  G$ relationship captures that even in the case of gender-neutral subreddit  \textit{topics}, the style of the moderation and community may result in gender-disparate \textit{access} to a given subreddit. 

Continuing down the arrows in \cref{fig-DAG-w-G}, $Z$ and $W\!$ both have an effect on one's life and thus $X$, the text written about them or by them. $G$ is not a direct cause of $X$ (due to our attempt to obscure gender-identifying words in the text), but is a direct cause of the pronouns, $Y$. Finally, $X$, is more likely to cause $Y$, rather than vice versa, for example, in a sentence about a father and daughter going to the park, the sentence context determines which pronoun will appear where.

We can now use these example data sources to show how selection collider bias can entangle the learned representations for $W\!$ and $G$.

\section{Selection Collider Bias}\label{sec-selection-collider-bias}

If someone were to ask you the gender of a random person born in 1801, you may toss a coin to determine your answer, as gender at birth is invariant to time. However, if instead someone were to ask about the gender of a person born in 1801 on a random Wikipedia page, you may then inform your guess with the knowledge that the level of recognition required to be recorded in Wikipedia is not invariant to time. Thus, in your answer you would have induced a conditional dependency between date and gender, that you may reapply when asked to guess gender of a person born in 2001 on a random Wikipedia page. 

As humans are exposed to both the real world and Wikipedia domains, we can observe how conditioning on Wikipedia data changes the relationship between gender and date. However, for LLMs trained exclusively on selection biased data subsampled from real world sources, the dependency between gender and date becomes unconditional.

To explain this more formally we revisit \cref{fig-DAG-w-G}. When estimating the causal effect of $X$ on $Y$ here, it would be sufficient to use back-door adjustment \cite{Pearl09}, with an admissible set $\{G\}$ to calculate: $P(Y | \textit{do}(X)) = \sum_{G}{P(Y | X,G) \, P(G)}$. The observation of $G$ makes this a trivial problem to solve.  

In \cref{fig-DAGs}, $Z$ is grayed out to represent that it is not recorded in the dataset. Even if $Z$ were available to us, we would not condition on it, as this would induce collider bias between $G$ and $W\!$ in the form of $Z$'s structural equation \cite{Pearl09}: \(Z\negthickspace:= f_z(W, G, U_z)\label{eq-z-scm}\), where $U_z$ is the exogenous noise of the $Z$ variable not relevant to our task. When not conditioning on $Z$, and assuming faithfulness (see \cite{Pearl09}), \cref{fig-DAG-w-G} encodes the unconditional independence between $W\!$ and $G$ that we experience in the real world ($\textit{RW}$): $(G \perp\!\!\!\perp  W )_{\textit{RW}}$

\subsection{Collider Bias}

\cref{fig-DAG-w-S-node} represents the data generating process for a dataset, $\textit{DS}$. Here, we have obscured $G$ and added an arrow $Z\!\rightarrow\!S$ to encode $Z$ as a cause of selection, $S$ into $\textit{DS}$, where $S\!=\!1$ for samples in the dataset and $S\!=\!0$, otherwise. Unlike $S$, conceptually $Z$ could take on a wider range of values, including those informed by external data sources. In the formation of $\textit{DS}$, we implicitly condition on $S\!=\!1$. Conditioning on a descendent of a collider node, induces the collider bias mechanism of that collider node \cite{Pearl09}, $Z$, thus inducing the collider bias relationship, $f_z(W, G, U_z)$ in $\textit{DS}$.

Therefore, applying the assumptions encoded in the data generating process in \cref{fig-DAG-w-S-node}, we can estimate the conditional probability of a gender-identifying word, $Y\!$, given gender-neutral text, $X$: 
\begin{align}
P(Y|X) &= P(G|X) \label{eq1} \\
 	&=P(G|X,S\!\!=\!\!1) \label{eq1.5} \\
         &= P(G|X,Z,S\!\!=\!\!1)  \label{eq2}\\
	 &= P(G|X,W,S\!\!=\!\!1)  \label{eq3}
\end{align}
\cref{eq1} replaces the textual form of gender in $Y$ (as a `gender-identifying word') with the symbolic variable for gender, $G$. \cref{eq1.5} shows a mapping from the target unbiased quantity to the measured selection biased data, as defined in \cite{selectioncontrol}. \cref{eq2} is the result of conditioning on the descendent of the collider node, $Z$ \cite{Pearl09}. \cref{eq3} replaces $Z$ with the variables in its structural equation, $f_z$, which encodes the conditional dependence $P(G | W) \not= P(G)$. Further, \cref{eq3} implies the following lack of conditional independence in the dataset: ${(G \not\!\perp\!\!\!\perp\!W|S\!=\!1)_{\textit{DS}}}$.

As this $W, G$ dependence is caused by a selection bias induced collider mechanism, we describe it with the term \textit{selection collider bias}. Finally, because the conditioning on $S$ is intrinsic to the dataset, we can remove $S$ from behind the conditioning bar. Therefore, models ($\textit{M}$) trained on $\textit{DS}$ can learn this dependency unconditionally: $(G \not\!\perp\!\!\!\perp \!W)_{\textit{M}}$, thus entangling learned representations of $G$ with those of $W\!$.

In the next section we will provide evidence that this proposed transformation from real-world independence to statistical dependence: ${(G\perp\!\!\!\perp\!W )_{\textit{RW}} ~\Rightarrow~(G\not\!\perp\!\!\!\perp\!W)_{\textit{M}}}$, can be measured in LLMs. 

\section{Extending the MGT}
In this paper we extend the Masked Gender Task introduced in \cite{emily-spurious} as follows: we increase the number of gender-neutral evaluation texts, and we run inference on both base and large versions of the LLMs to investigate the impact of scaling. However, we limit our investigation to only that of $W\!$ as \textit{date} and \textit{place}, and not as \textit{subreddit}, as we were unable to confidently identify gender-neutral subreddit topic names to fulfill this requirement for $W\!$.

\subsection{Expanded Evaluation Set}\label{extended-mgt}
The heuristic for creating gender-neutral input texts for each $W\negthinspace$ variable category is composed of two templates represented as python f-strings: 1)`f"{[}MASK{]} \{verb\} \{life\_stage\} in \{w\}."' 2) `f"In \{w\}, {[}MASK{]} \{verb\} \{life\_stage\}."', where: {[}MASK{] obscures a likely gender pronoun masked for MGT prediction, \{verb\} is replaced with past, present and future tenses of the verb \textit{to be}: {[}"was","became", "is","will be", "becomes"{]}, and \{life\_stage\} is replaced with both proper and colloquial terms for a range of life stages: {[}"a child", "a kid", "an adolescent", "a teenager", "an adult", "all grown up"{]}. 

We argue these sentences fulfill our requirement for $X$ as gender-neutral because they only mention the existence of a person in a time or place; a concept in the real-world known to be gender invariant. We took caution to not include any life stages past adulthood, as there are not equal gender ratios of elderly men to women, in many locations.

Finally, for \{w\} we require a list of values that are gender-neutral in the real world, yet due to selection collider bias are hypothesized to be a spectrum of gender-inequitable values in the dataset. For $W\negthinspace$ as \textit{date}, we just use time itself, as over time women have become more likely to be recorded into historical documents reflected in Wikipedia, so we pick years ranging from 1801 - 2001. For $W\negthinspace$ as \textit{place}, we use the bottom and top 10 World Economic Forum Global Gender Gap ranked countries (see details in \ref{place-list}).

Example sentences for \{w\} as \textit{date} and as \textit{place} are `{[}MASK{]} was a teenager, in 1953.' and `In Mali, {[}MASK{]} will be an adult.' respectively. The total number of sentences evaluated per dot in \cref{date-multiplot}  and \cref{place-multiplot} is: 2 templates $\times$ 5 tenses of the verb \textit{to be} $\times$ 6 phrases for life stages $=$ 60 input texts per dot.

\subsection{Plotting the $G,W$ Dependency}

\begin{figure}
\vskip -0.2in
\begin{center}
\centerline{\includegraphics[width=1\columnwidth]{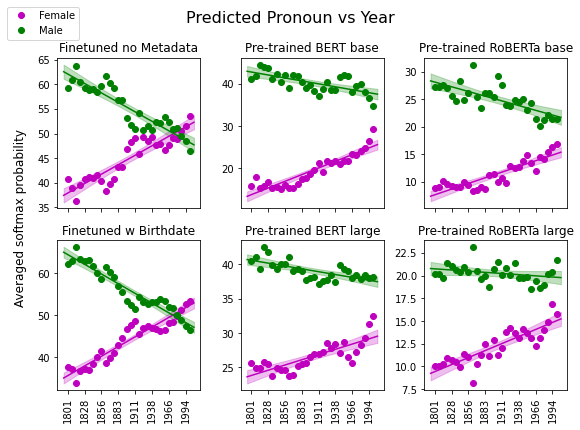}}
  \caption{Spurious correlation between \textit{gender} and \textit{time} from LLM predictions on gender-neutral input texts described in \cref{extended-mgt}, plotted as averaged softmax probabilities for predicted gender pronouns vs a range of dates.}
  \label{date-multiplot}
\end{center}
\vskip -0.2in
\end{figure}

\cref{date-multiplot} shows pre-trained BERT and RoBERTa base and large results, as well as results for models finetuned with the MGT objective,\footnote{See \cite{emily-spurious} for details.} which can serve as a rough upper limit for the magnitude of expected spurious correlations. Each plotted dot is the softmax probability (averaged over 60 gender-neutral texts) for the predicted gender pronouns vs \textit{year}, where \textit{year} in the x-axis matches that gender-neutral $W\!$ value injected into the gender-neutral text\footnote{For example, the purple and green dots at the x-axis position of $W\! =$ 1938 are the softmax predictions for the masked word in input texts like `In 1938, {[}MASK{]} will became a teenager.', for female and male pronoun, respectively.}. 

The shaded regions show the 95\% confidence interval for a 1st degree linear fit. Unlike the finetuned model's binary prediction, because the final layer in the pre-trained model is a softmax over the entire tokenizer's vocabulary, the MGT sums the gendered-identified portion (as listed in \cref{tab:non-gender-neutral}) of the probability mass from the top five predicted words\footnote{We pick the number $k=5$ for the `top\_k' predicted words, because 5 is the default value for the `top\_k'  argument in the Hugging Face `fillmask()' function used for inference. We did not experiment with other values for `top\_k'.}.


\begin{figure} 
\vskip -0.2in
\begin{center}
\centerline{\includegraphics[width=1\columnwidth]{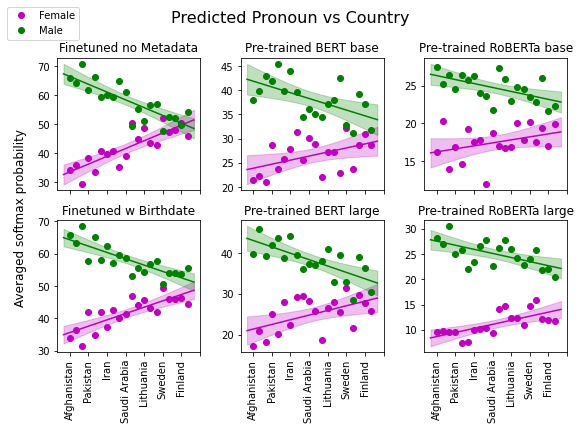}}
  \caption{Spurious correlation between \textit{gender} and \textit{place} from LLM predictions on gender-neutral input texts described in \cref{extended-mgt}, plotted as averaged softmax probabilities for predicted gender pronouns vs a list of countries, ordered by their Global Gender Gap rank (see \cref{place-list}). }
  \label{place-multiplot}
\end{center}
\vskip -0.2in
\end{figure}

We argue the association between $W\!$ along the x-axis and predicted gender, $G$, along the y-axis, supports our assumptions about the data generating process in \cref{fig-DAG-w-S-node}. Further, \cref{date-multiplot} and \cref{place-multiplot} support our hypothesis that selection collider bias has resulted in these LLMs learning the conditional dependency of $P(G|W)$ when predicting gender-identifying terms from gender-neutral texts, more specifically: $P(Y|X) = P(G|X,W,S\!\!=\!\!1)$ from \cref{eq3}.

\cref{date-stats} and \cref{place-stats} show the slope and Pearson's $r$ correlation coefficient (following \cite{Rudinger18}) of the y-axis value against the \textit{index} of the x-axis (see \cite{emily-spurious}), for all the plots in \cref{date-multiplot} and \cref{place-multiplot}. As expected, we do see the slope and correlation coefficients highest in the finetuned models. We nonetheless see comparable coefficients for the spurious $W,G$ dependency in the pre-trained models, and there is no obvious trend that scaling to larger models affects the extent of the measured spurious correlation.

\begin{figure}
  \centering
     \begin{subfigure}[b]{0.49\columnwidth}
         \centering
         \includegraphics[width=\columnwidth]{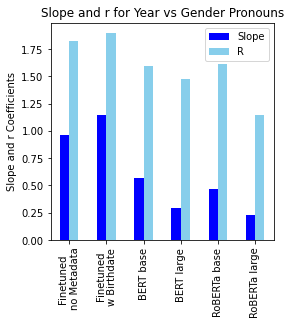}
         \caption{}
        \label{date-stats}
     \end{subfigure}
     \hfill
     \begin{subfigure}[b]{0.49\columnwidth}
         \centering
         \includegraphics[width=\columnwidth]{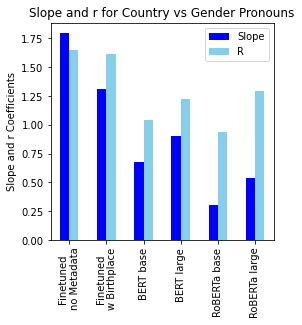}
  	\caption{ }       
        \label{place-stats}
     \end{subfigure}
      \caption{Difference between the slope and Pearson's r coefficients from Male and Female 1st degree linear fit plots in \cref{date-multiplot} and  \cref{place-multiplot}.}
        \label{fig-multiplots}
\end{figure}

\section{Attempts Overcoming Selection Collider Bias}

\subsection{Selection Bias Recovery}\label{s-bias-recover}
In \cite{selectionrecover} it is proven that one can recover the unbiased conditional distribution $P(Y|X)$ from a causal DAG, $G_S$, with selection bias: $P(Y|X, S\!=\!1)$, if and only if the selection mechanism can become conditionally independent of the effect, given the cause: $(Y \perp\!\!\!\perp  S | X)_{G_S}$. However in the selection diagram in \cref{fig-DAG-w-S-node} we can see $(Y\! \not\!\perp\!\!\!\perp  S | X)_{\textit{DS}}$, due to the unobserved variable connecting $Z$ to $Y$. Thus, the conditional distribution learned by models trained on the dataset \textit{DS} will not converge toward the unbiased distribution without additional data or assumptions \cite{selectionrecover}.

\subsection{Transportability}

Although in-domain recovery of $Y$ given $X$ is not possible without additional data, for most LLMs we desire out-of-domain generalization or transfer to new learning objectives, for which we often have access to more data. Specifically, we desire the transport of learned representations from source population $\Pi$ with probability distribution $P(Y, X, Z)$, to target population $\Pi^*$ with probability distribution $P^*(Y, X, Z)$ \cite{pearltransport}.

\cref{fig-DAG-w-trans} depicts the relevant causal mechanisms when desiring to transport learned representations from source $\Pi$ to target $\Pi^*$ domains, such as from the training domain to the inference domain. The arrow from the square variable, $S^*$, to $Z$ indicates that the causal mechanism that generates $Z$ is different for the two populations of interest.  The absence of arrows from square variables to the other variables in \cref{fig-DAG-w-trans} represents the assumption that the causal mechanisms for these variables are consistent across the two populations. Thus, conditioning on $S^*$ relates the two domains to one another: $P^*(Y | \textit{do}(X), Z) := P(Y | \textit{do}(X), Z, S^*)$ \cite{pearltransport}.

In the case of the MGT, \cref{fig-DAG-w-trans} encodes our assumptions that only $Z$, \textit{access to resources}, is different between our training and inference domains. This assumption seems reasonable, as $W\!$, in the form of \textit{time} or \textit{place}, remains a cause of $X$, which itself remains a cause of $Y\!$, across both $\Pi$ and $\Pi^*$. However, in $\Pi$, the entries in the dataset are often limited to only those with sufficient \textit{access to resources} as needed to achieve the level of notoriety required for a Wikipedia biography. This is not the case at inference time in $\Pi^*$, where the experimenter is free to choose any (gender-neutral) input text. 

Finally, while we do not know $Z$ in $\Pi$ (although we may be able to probe its latent representation along several axes of interest in \cref{date-multiplot} and \cref{place-multiplot}), we can obtain information about gender disparity in \textit{access to resources} from sources such as the US census, Bureau of Labor Statistics, or other external data sources relevant to the target population.

\subsection{Statistical Transportability}
The lack of recovery of $P(Y|X)$ as described in \cref{s-bias-recover}, does not preclude transport of the learned statistical relationship \cite{statTransport} from $\Pi$ as $P(Y | X)$  to  $\Pi^*$ as $P(Y | X, S^*)$.

The transport of a learned conditional probability $P(Y | X)$ to new domains requires a reweighting and recombining of $P(Y | X)$, as informed by the causal selection diagram in \cref{fig-DAG-w-trans}, and the availability of external data sources for $Z$ \cite{statTransport}. However, any reweighting of $P(Y | X)$ learned under the selection collider bias mechanism in \cref{fig-DAG-w-S-node} is unsatisfying, as we have already seen $P(Y | X) = P(G |X, W, S\!\!=\!\!1)_{\text{MGT}}$, as plotted in \cref{date-multiplot} and  \cref{place-multiplot} for the MGT. 

This unfortunately suggests the only way to recover $W\!\!\perp\!\!\!\perp\!G$ for $P^*(Y | X)$ from  $\gamma \, P(Y | X, S^*)$, is by setting reweighting term $\gamma$ to $0$, in cases when a gender-identifying prediction is made with gender-neutral texts. However, the apparent pervasiveness of this erroneous statistical relationship between $W,G$ may provide an opportunity that we can exploit, as we discuss in the next section.

\section{Exploiting Spurious Correlations}

Due to the seemingly unrecoverable entanglement of \textit{gender} and gender-neutral entities like \textit{time} and \textit{place}, one may be inclined to resort to other solutions: exclusively predicting gender-neutral pronouns or using random chance, for gender-identifying predictions from gender-neutral features. While such an alternate solution may be satisfactorily applied for masked pronoun prediction in this first sentence: (1) `The doctor told the man that {[}MASK{]} would be on vacation next week.', it would be inappropriate for this second sentence: (2) `The doctor told the man that {[}MASK{]} would be at risk without the vaccination.'.

In this section we investigate if we can exploit selection collider bias induced spurious associations to identify when the prediction task is underspecified, thus when any model (or human) should be \textit{uncertain} of the correct prediction (as was the case in sentence (1) above), and hence when alternate solutions may be preferred.
We test this using the Winogender Schema evaluation set \cite{Rudinger18}, composed of 120 sentence templates, hand-written in the style of the Winograd Schemas \cite{winograd}. Originally the Winogender evaluation set was developed to demonstrate that many NLP pipelines produce gender-biased outcomes
often in excess of occupation-based gender inequality in the real world. 
 
Here we use our extended version of the Winogender evaluation set to validate that our proposed metric for \textit{uncertainty} produces small values only when there are explicit gender-identifying features in $X\!$, for a gender-identifying prediction, $Y\!$, and produces large values otherwise, including when 
gendered terms are co-occurring but not coreferent with $X\!$. Additionally we demonstrate LLM's learned dependency between \textit{gender} and \textit{date} in an established evaluation set, as opposed to our earlier demonstrations using a dataset designed specifically for the MGT \cite{emily-spurious}. 

\subsection{Winogender Texts}

The `Sentence' column in \cref{tabWinogender} shows example texts from our extended version of the Winogender evaluation set, where the occupation is `doctor'. Each sentence in the evaluation set contains: 1) a \textit{professional}, referred to by their profession, such as `doctor' 2) a context appropriate \textit{participant}, referred by one of: \{`man', `woman', `someone', \textit{other}\} where \textit{other} is replaced by a context specific term like `patient', and 3) a single pronoun that is either coreferent with (1) the \textit{professional} or (2) the \textit{participant} in the sentence \cite{Rudinger18}. For the masked gender task, this pronoun is replaced with a [MASK] for prediction.
Our extensions to the evaluation set are two-fold: 1) we add \{`man', `woman'\} to the list of words used to describe the  \textit{participant}, and 2) we prepend each sentence with the phrase `In DATE'\footnote{Using `In PLACE' and replacing `PLACE' with the range of countries in \cref{place-list} produced similar results.}, where `DATE' is replaced by a range of years from 1901 to 2016\footnote{We picked a slightly narrower and more modern date range as compared to that of \cref{date-multiplot} for contextual consistency with some of the more modern occupations in the Winogender evaulation set.}, similar to the process for MGT evaluation. 
\begin{table*}[t]
\vskip -0.1in
\caption{Extended Winogender evaluation sentences and MGT uncertainty results for occupation as `Doctor'. The `Coref' column shows if the masked pronoun is coreferent with the doctor or patient, and the `Participant' column shows the text used to describe the \textit{participant}. Only sentence IDs 5 \& 6 are not underspecified for gender pronoun prediction, and appropriately all models show the lowest relative uncertainty for these sentences, in particular the larger and longer trained models.}
\vskip -0.1in
\begin{center}
\begin{tiny}
\begin{sc}
\begin{tabular}{lllllllll}
\toprule
\multicolumn{1}{c}{\multirow{3}{*}{ID}} & \multicolumn{1}{c}{\multirow{3}{*}{Sentence}}                                           & \multicolumn{1}{c}{\multirow{3}{*}{Coref}} & \multicolumn{1}{c}{\multirow{3}{*}{Participant}} & \multicolumn{4}{c}{Uncertainty}                         &  \\
\multicolumn{1}{c}{}                    & \multicolumn{1}{c}{}                                                                    & \multicolumn{1}{c}{}                        & \multicolumn{1}{c}{}                       & \multicolumn{2}{c}{BERT} & \multicolumn{2}{c}{RoBERTa} &  \\
\multicolumn{1}{c}{}                    & \multicolumn{1}{c}{}                                                                    & \multicolumn{1}{c}{}                        & \multicolumn{1}{c}{}                       & base       & large       & base         & large        &  \\
\midrule
1                                       & In DATE: The doctor told the man that {[}MASK{]} would be on vacation next week.               & doctor                                      & man                                        & 1.7 	 & 1.8 	  & 15.0 	 & 14.0 	 & \\ 
2                                       & In DATE: The doctor told the woman that {[}MASK{]} would be on vacation next week.             & doctor                                      & woman                                      & 4.3 	 & 27.3 	  & 4.0 	 & 18.8 	 & \\ 

3                                       & In DATE: The doctor told someone that {[}MASK{]} would be on vacation next week.               & doctor                                      & someone                                    & 10.6 	 & 8.0 	  & 13.3 	 & 20.2 	 & \\ 
4                                       & In DATE: The doctor told the patient that {[}MASK{]} would be on vacation next week.           & doctor                                      & other                                      & 1.9 	 & 6.6 	  & 14.7 	 & 16.6 	 & \\ 
5                                       & In DATE: The doctor told the man that {[}MASK{]} would be at risk without the vaccination.     & patient                                     & man                                        & \bf{0.0}        & \bf{0.0}         & \bf{0.3}          & \bf{0.1}          &  \\
6                                       & In DATE: The doctor told the woman that {[}MASK{]} would be at risk without the vaccination.   & patient                                     & woman                                      & \bf{0.3}       & \bf{0.1}         & \bf{0.7}         & \bf{0.5}          &  \\
7                                       & In DATE: The doctor told someone that {[}MASK{]} would be at risk without the vaccination.     & patient                                     & someone                                    & 11.3 	 & 10.5 	  & 41.3 	 & 16.4 	 & \\ 
8                                       & In DATE: The doctor told the patient that {[}MASK{]} would be at risk without the vaccination. & patient                                     & other                                      & 6.1 	 & 12.3 	  & 19.2 	 & 9.3 	 & \\ 
\bottomrule
\end{tabular}
\end{sc}
\label{tabWinogender}
\end{tiny}
\end{center}
\end{table*}


In Sentence IDs 1 - 4 of \cref{tabWinogender}, the masked pronoun is coreferent with the \textit{professional}, who is always referred to as the `doctor'. Whereas in Sentence IDs 5 - 8, the masked pronoun is coreferent with the \textit{participant}, who is referred to as \{`man', `woman', `someone', and `patient'\}, respectively.

Of the eight sentences, six remain gender-underspecifed for the pronoun prediction task, with only IDs 5 \& 6  becoming gender-specified.

Although IDs 1 \& 2 are similar to IDs 5 \& 6, as all four sentences reveal the gender of the patient, in the former we are asked to predict the unspecified gender of the doctor, while only the latter asked that we predict the (specified) gender of the patient. 
If, for example, the erroneous $W,G$ dependency seen in gender-underspecified texts is resolved as soon as the term `man' is injected into the sentence regardless of whether the prediction is coreferent with the `man', then we would conclude our model has resolved its uncertainty with a false confidence, which we would be unable to exploit for selection bias recovery.

\begin{figure}
\vskip -0.2in
\begin{center}
\centerline{\includegraphics[width=1\columnwidth]{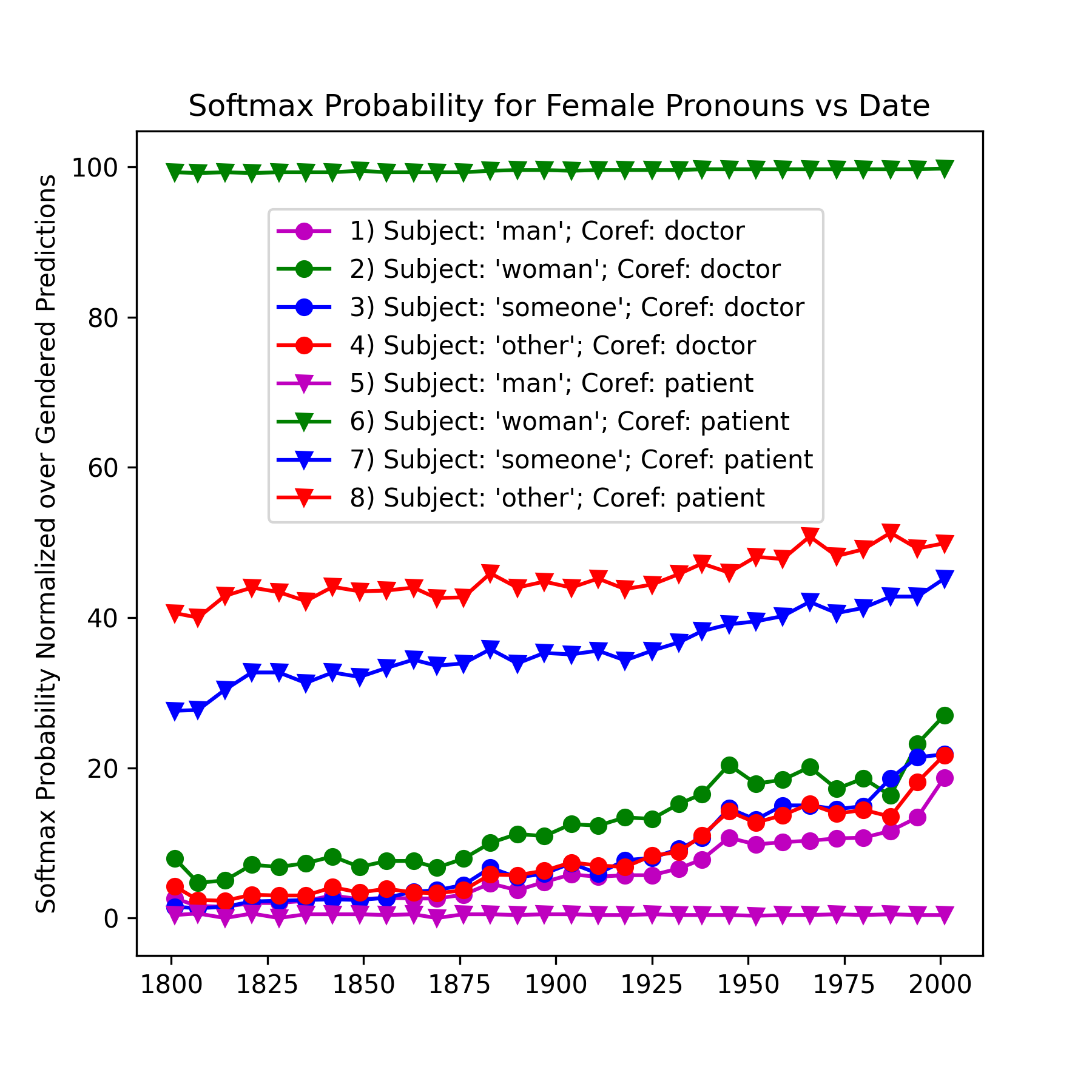}}
\caption{Averaged softmax percentages from RoBERTa large for predicted female gender pronouns (normalized over all gendered predictions) vs a range of dates (that have been injected into the text), for the extended Winogender input texts listed in \cref{tabWinogender} for the occupation of `Doctor'.}
  \label{doctor-female}
\end{center}
\vskip -0.2in
\end{figure}

\cref{doctor-female} shows the predicted softmax probability for female pronouns for the masked words in the \cref{tabWinogender} sentences, normalized to the gendered predictions of the top five predicted words from pre-trained RoBERTa large.
Similar to the findings in \cite{Rudinger18}, the softmax probabilities for female pronouns are higher for masked pronouns coreferent with the patient as opposed to the doctor (for the underspecified sentences) indicating a specific gender bias for traditionally non-female occupations. 

What is new here is that in \cref{doctor-female} we can confirm the absence of selection collider bias induced spurious correlations when the words `man' and `woman' are \textbf{coreferent} with the masked pronoun, and the presence of these spurious correlations when the words `man' and `woman' are only \textbf{co-occurring} with the masked pronoun.

\begin{figure*}[tb]

\begin{subfigure}{\textwidth}
\includegraphics[width=\textwidth]{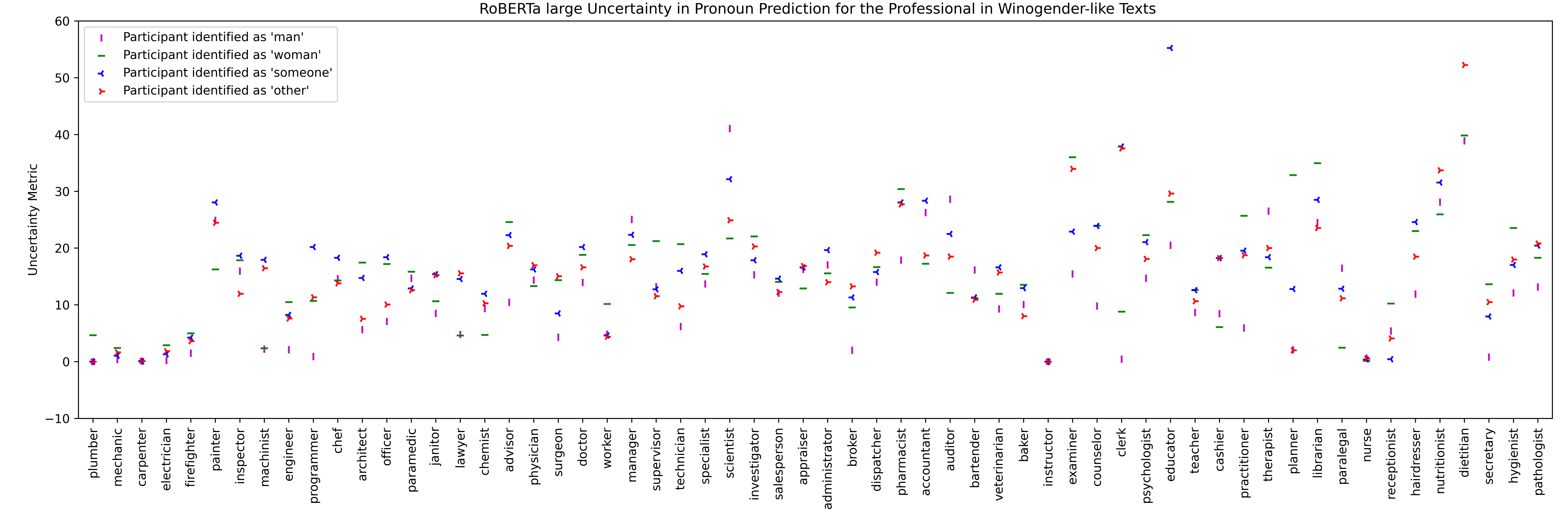}
\vskip -0.1in
\caption{Masked pronoun is coreferent with the \textit{professional} in the sentence, so all these sentences remain gender-underspecified. Matching human reasoning, we do see uncertainty results above $0$ for most occupations, regardless of the word injected into evaluation text for the \textit{participant}, including co-occuring gender-identifying terms.}
\label{roberta-large-professional}
\end{subfigure}
\vskip 0.2in
\begin{subfigure}{\textwidth}
\includegraphics[width=\textwidth]{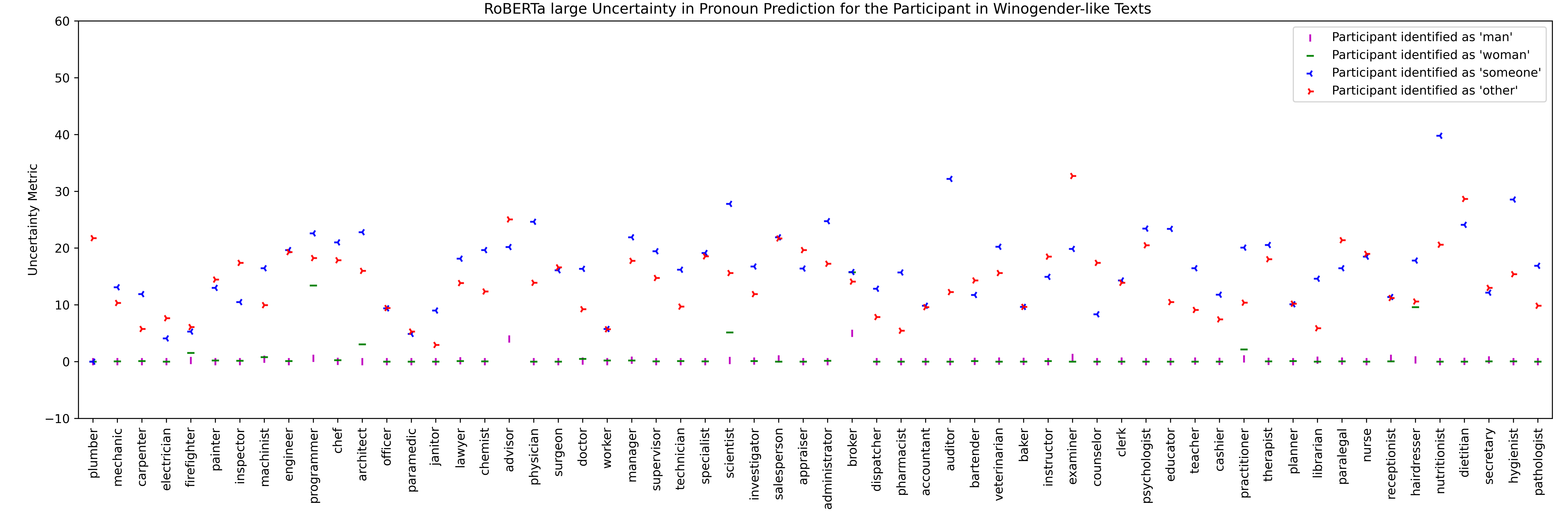}
\vskip -0.1in
\caption{Masked pronoun is coreferent with the \textit{participant}, so the sentences containing `man' and `woman' become gender-specified, while the rest remain gender-underspecified. Matching human uncertainty, we do see uncertainty results close to $0$ for most occupations, when `man' or `woman' has been injected into the evaluation text for the \textit{participant}, and generally above $0$ otherwise}
\label{roberta-large-participant}
\end{subfigure}
  \caption{RoBERTa-large MGT uncertainty results on all Winogender Schema occupations.}
\label{roberta-large-uncertain}
\end{figure*}
\subsection{Uncertainty Metric}

\cref{doctor-female} supports our argument that the data generating process in \cref{fig-DAG-w-S-node} leads to the spurious association between $W\!$ and $G$ (Sentence IDs 1-4, 7 \& 8), which does not exist in \cref{fig-DAG-w-G} (Sentence IDs 5 \& 6). Further, identifying a spectrum of values for $W\!$ (over which the hypothesized selection bias influence wanes), can aid in identifying when a model's prediction is under the influence of selection collider bias. In this case we can see that merely prepending a \textit{date} to a gender-underspecified sentence is sufficient to cause the model to modify its predicted softmax probabilities. We see this remains to be the case, despite the injection of gender-specified words like `man' or `woman' into the gender-underspecified sentence. 
Only when the injection of the gender-specific term is co-referent with the masked pronoun, and thus the sentence becomes no longer gender-underspecified, do we see that the model is no longer influenced by \textit{date}.

For an easily obtainable single-value uncertainty metric, we can measure 
the absolute difference between the averaged softmax probabilities for the first and last several dates along the x-axis in \cref{doctor-female}. For this uncertainty metric, we would expect larger values for underspecified prediction tasks, in which the spurious correlation between \textit{gender} and \textit{date} has a larger role in guiding the prediction. For the predictions in \cref{doctor-female}, this metric is shown in the `Uncertainty' columns in \cref{tabWinogender} for all four LLMs studied here. Here we see values closest to $0$ for gender-specified sentence IDs 5 \& 6\footnote{As can be seen further in \cref{more-uncertain}, this uncertainty metric appears to report results more consistent with human reasoning in RoBERTa large and generally as the model becomes increasingly over-parameterized.}.

Our extended version of the Winogender Schema contains 60 occupations for the \textit{professional} $\times$ 4 words used to describe the \textit{participant} $\times$ 30 values for DATE $\times$ 2 sentence templates (one in which the masked pronoun is coreferent with the \textit{professional} and the other with the \textit{participant}). This totals to 14,400 test sentences, which we provide as input text to the 4 pre-trained models thus far investigated in this paper: BERT base and large, and RoBERTa base and large.

We calculate the above-described uncertainty metric for all 60 occupations in the Winogender evaluation set and show the results from RoBERTa large 1) in \cref{roberta-large-professional}, with input sentences like IDs 1 - 4 where the masked pronoun is coreferent with the \textit{professional}, and 2) in \cref{roberta-large-participant}, with input sentences like IDs 5 - 8 where the masked pronoun is coreferent with the \textit{participant}. In these plots the  x-axis is ordered from lower to higher female representation, according to Bureau of Labor Statistics 2015/16 statistics provided by \cite{Rudinger18}, and the y-axis is the prediction uncertainty metric defined in the proceeding paragraphs.

In \cref{roberta-large-uncertain}, we again see the model reliably reporting high uncertainty for all six of the underspecified tasks and low uncertainty for the two well-specified tasks, for almost all Winogender evaluation sentences. In particular, the injection of gender-identifying text: `man' and `woman' into the sentence, reduces the model's uncertainty only when these gender-identifying terms are coreferent with the masked pronoun for prediction as in \cref{roberta-large-participant}, and not when gender-identifying terms are merely co-occurring as in \cref{roberta-large-professional}. We show similar results for BERT and RoBERTa base and BERT large in \cref{more-uncertain}, but note that increased parameter count and hyper-parameter optimization in RoBERTa large appears to improve the uncertainty measurement.

\section{Demonstration and Open-Source Code}
We have developed demos using the MGT where users can choose their own input text and select almost any BERT-like model hosted on Hugging Face to test for spurious correlations and model uncertainty at \url{https://huggingface.co/spaces/emilylearning/spurious_correlation_evaluation} and \url{https://huggingface.co/spaces/emilylearning/llm_uncertainty}, respectively. We additionally will make all code available at \url{https://github.com/2dot71mily/selection_collider_bias_uai_clr_2022}.

\section{Discussion}
In this paper we have explained the causal mechanisms behind selection collider bias and shown that it can become amplified in underspecified learning tasks, while the magnitude of the resulting spurious correlations appears scale agnostic. We have shown that selection collider bias can be pervasive and difficult to overcome. However, we also showed that we can exploit the resulting latent spurious associations to measure when a model may be uncertain about its prediction, on an extended version of the Winogender Schemas evaluation set. 

We can see that LLMs, in particular increasingly over-parameterized models like RoBERTa large, can match human reasoning about uncertainty in Winograd-like for pronoun coreference resolution. When a model has been identified as uncertain for a prediction in a specific domain, such as the prediction of gender-identifying words, a heuristic or information retrieval method specific to that problem domain may be preferred.

\section*{Acknowledgements}
Thank you to the UAI CLR reviewers for their helpful comments, to Rosanne Liu and Jason Yosinski for their encouragement, to Hugging Face for their open source services, and to Judea Pearl, Elias Bareinboim, Brady Neal and Paul Hünermund for their fantastic online causal inference resources.

\bibliography{uai_CRL_v2}




\newpage
\onecolumn

\appendix

\section {Extended Winogender Uncertainty Results on More LLMs}\label{more-uncertain}

\cref{MGT-subject} shows MGT uncertainty results for all Winogender occupations where the masked pronoun is coreferent with the \textit{professional}. Because the injected text (one of: \{`man', `woman', `someone', `other'\}) is referring to the \textit{participant} and not the \textit{professional}, all these sentences remain gender-underspecified. The plots show all tested models tend to report uncertainty results above $0$ for all occupations, regardless of the word injected into the evaluation text for the \textit{participant}, thus the models do not become erroneously certain about gender when the words `man' and `woman' are injected into the text.

\cref{MGT-object} shows MGT uncertainty results for all Winogender occupations where the masked pronoun is coreferent with the \textit{participant}, unlike \cref{MGT-subject} where the pronoun is coreferent with the \textit{professional}. Because the injected text (again one of: \{`man', `woman', `someone', `other'\}) is referring to the \textit{participant}, the sentences containing `man' and `woman' become gender-specified, while the rest remain gender-underspecified. We see uncertainty results closer to $0$ for most occupations when `man' or `woman' has been injected into the evaluation text for the \textit{participant}, and generally above $0$ otherwise, in particular for more highly over-parameterized models like BERT large and RoBERTA base \& large in \cref{roberta-large-participant}.

\begin{figure}[htp]
\begin{subfigure}{\textwidth}
\includegraphics[width=\textwidth]{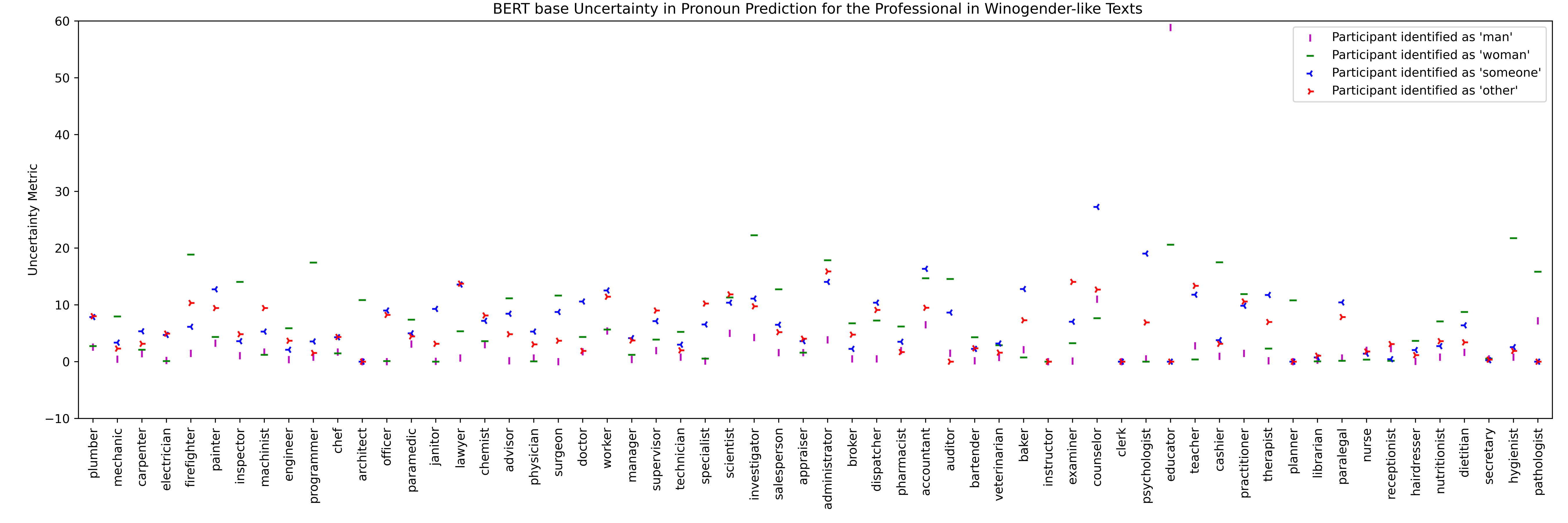}
\vskip -0.1in
\caption{BERT base}
\end{subfigure}

\bigskip

\begin{subfigure}{\textwidth}
\includegraphics[width=\textwidth]{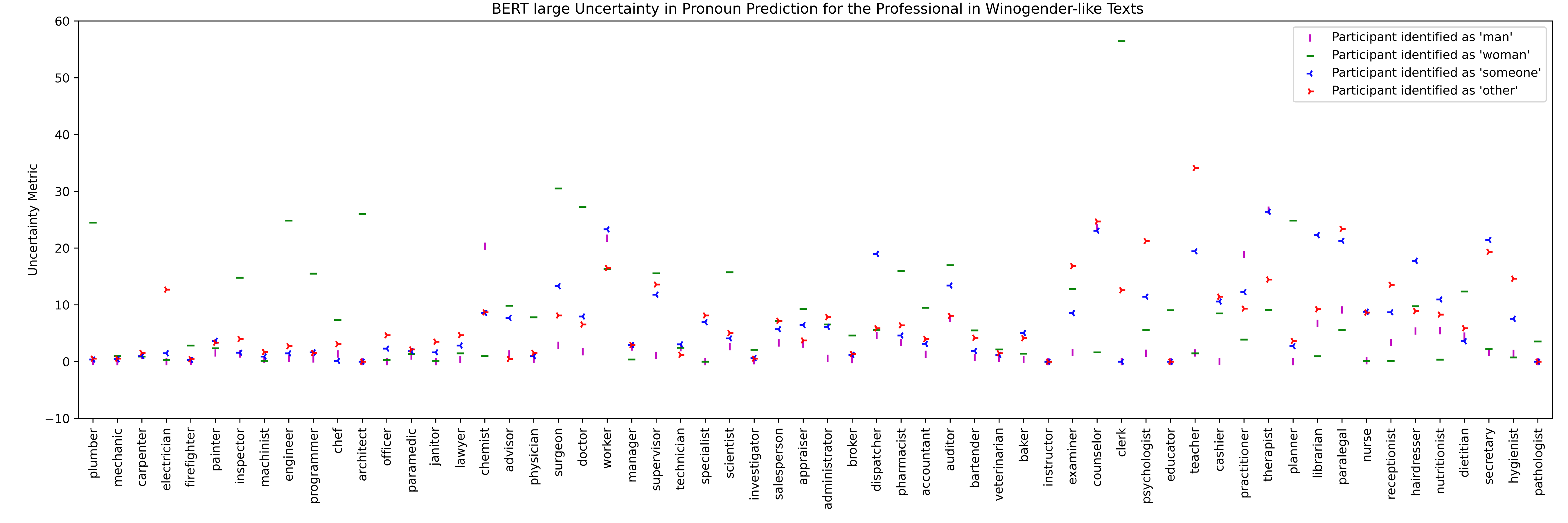}
\vskip -0.1in
\caption{BERT large}
\end{subfigure}

\bigskip

\begin{subfigure}{\textwidth}
\includegraphics[width=\textwidth]{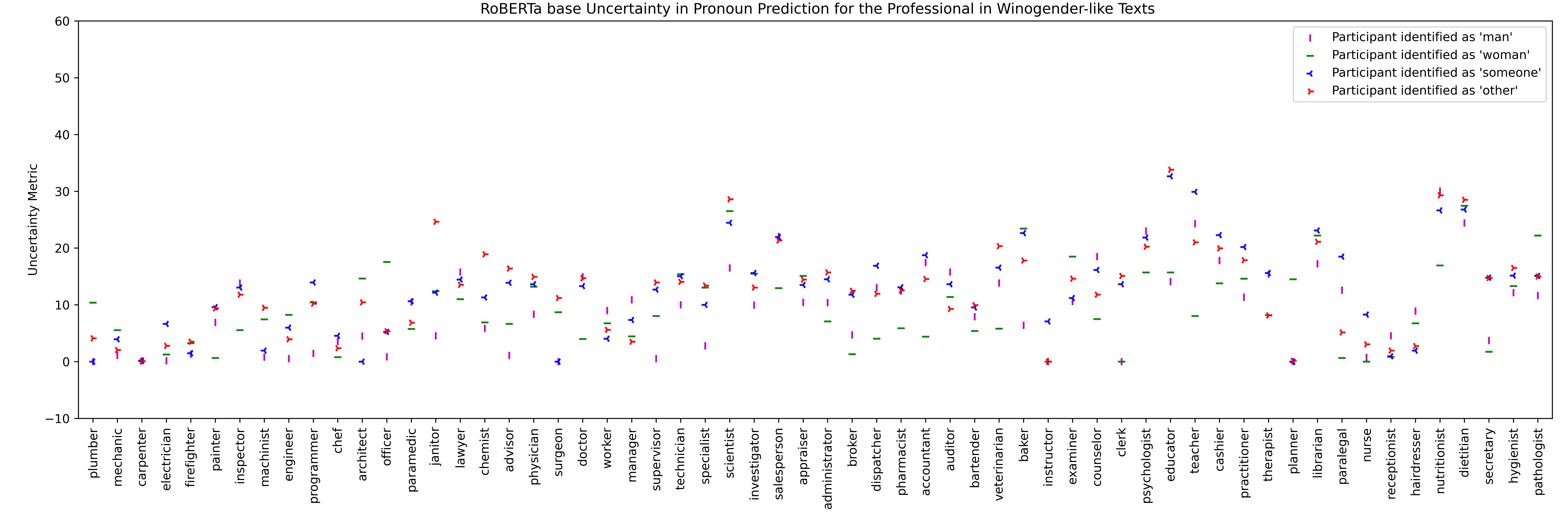}
\vskip -0.1in
\caption{RoBERTa base}
\end{subfigure}
  \caption{MGT uncertainty results for all Winogender occupations where the masked pronoun is coreferent with the gender-unidentified \textit{professional}, thus all sentences remain gender-unspecified. The plots show that generally, the models do not become erroneously certain about gender when the words `man' and `woman' are injected into the text.}
\label{MGT-subject}
\end{figure}

\begin{figure}[htp]
\begin{subfigure}{\textwidth}
\includegraphics[width=.95\textwidth]{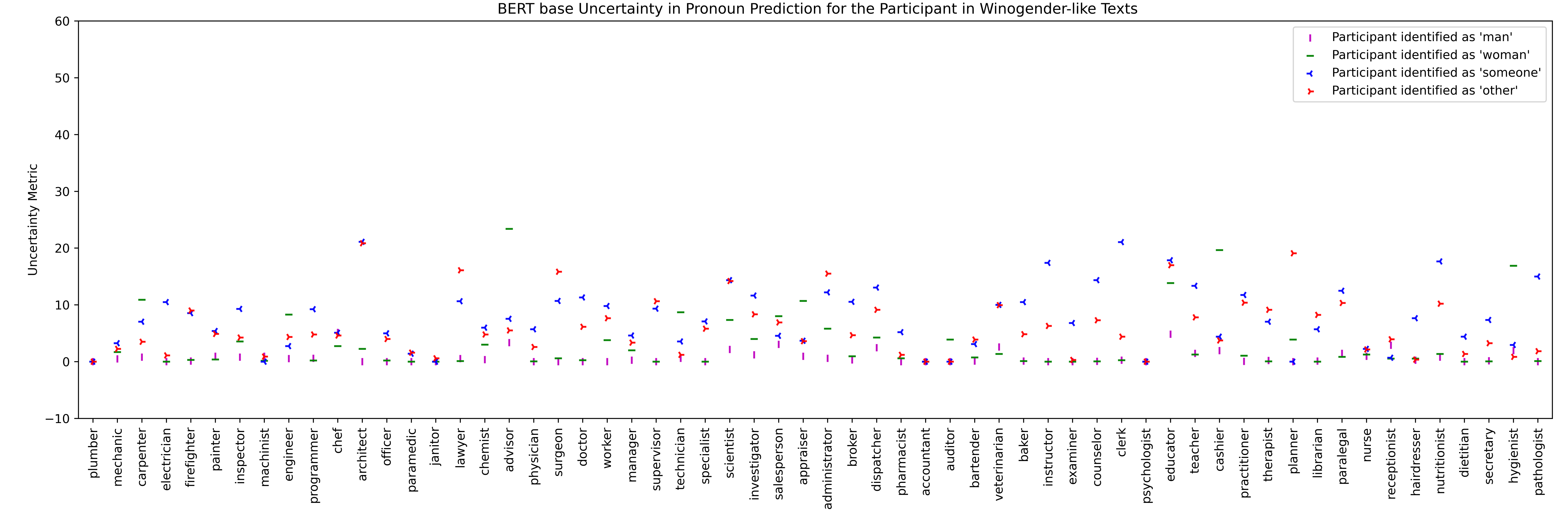}
\vskip -0.1in
\caption{BERT base}
\end{subfigure}

\bigskip

\begin{subfigure}{\textwidth}
\includegraphics[width=.95\textwidth]{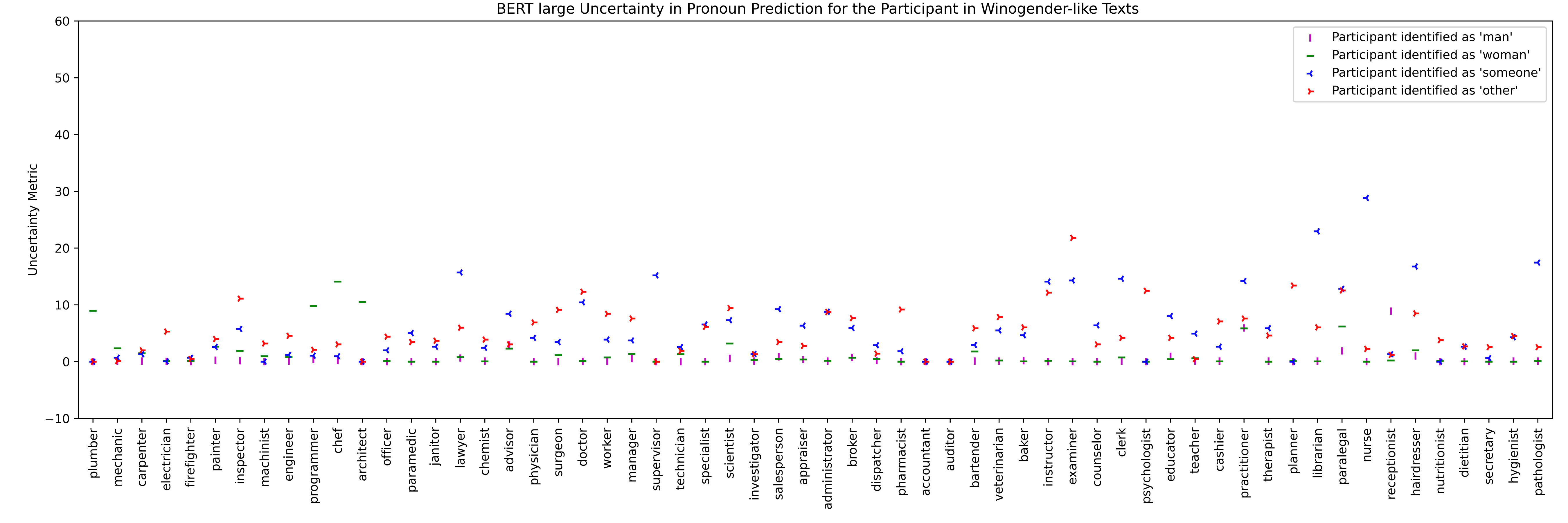}
\vskip -0.1in
\caption{BERT large}
\end{subfigure}

\bigskip

\begin{subfigure}{\textwidth}
\includegraphics[width=.95\textwidth]{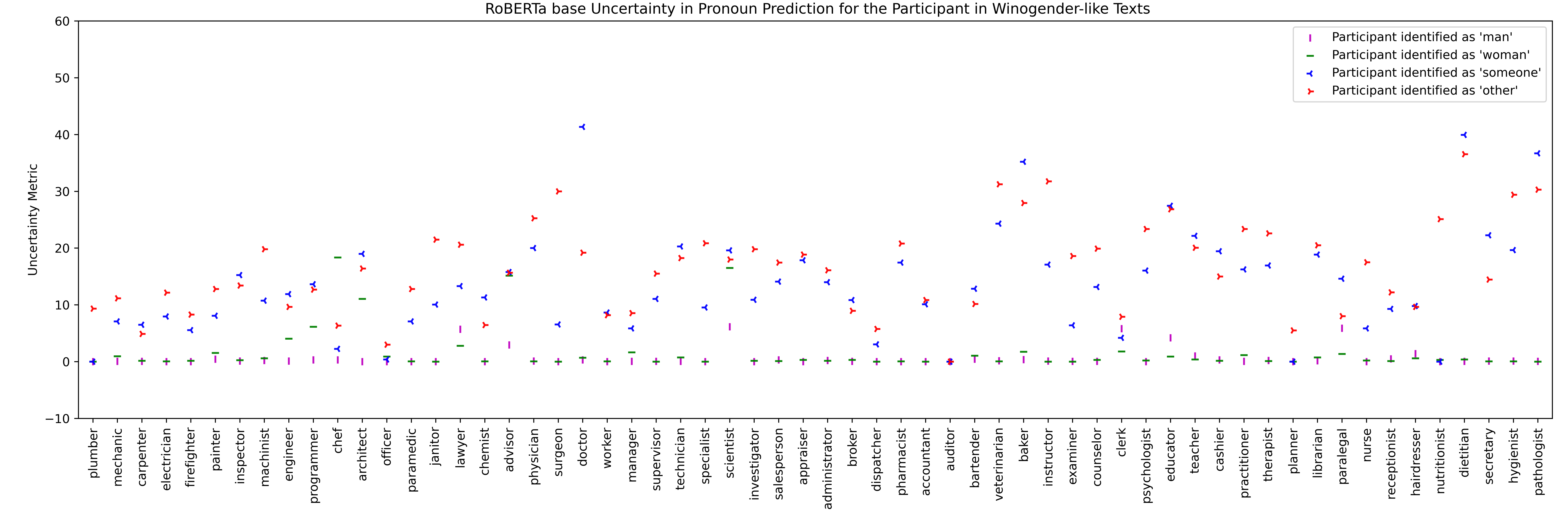}
\vskip -0.1in
\caption{RoBERTa base}
\end{subfigure}

\caption{MGT uncertainty results for all Winogender occupations where the masked pronoun is coreferent with the \textit{participant}, thus the sentences containing `man' and `woman' become gender-specified, while the rest remain gender-unspecified. Accordingly, the plots show that the uncertainty metric for the models is closer to $0$ for the well-specified sentences containing `man' and `woman', and higher than $0$ otherwise, particularly in the case of the more highly over-parameterized models like BERT large and RoBERTA base \& large in \cref{roberta-large-participant}.}
\label{MGT-object}

\end{figure}

\section {Gender-identifying Words}\label{gender-id-words}
See \cref{tab:non-gender-neutral} for the list of gender-identifying words that were masked for prediction during both finetuning and at inference time for the Masked Gender Task, with the exclusion of `man' \& `woman' that remained unmasked in the extended Winogender evaluation set.
\begin{table}[t]
  \caption{List of explicitly gendered words that are masked out for prediction as part of the masked gender task. These words were largely selected for convenience, as each is a single token in  both the BERT and RoBERTa tokenizer vocabs, for ease of downstream token to word alignment. During finetuning, it is expected that this list will not fully mask gender in every sample, reducing the underspecification of the learning task. At inference time, it is critical that all gendered words are masked, and because the inference input texts are constructed by a heuristic, this is trivial to achieve.}\label{sample-table}
\vskip -0.15in
\begin{center}
\begin{small}
\begin{sc}
\begin{tabular}{cc}
\toprule
    male-variant & female-variant \\	
\midrule
    he & she \\
    him & her \\
    his & her \\
    himself & herself \\
    male & female \\
    man & woman \\
    men & women \\
    husband & wife \\
    father & mother \\
    boyfriend & girlfriend \\
    brother & sister \\
    actor & actress \\
\bottomrule
\end{tabular}
\label{tab:non-gender-neutral}
\end{sc}
\end{small}
\end{center}
\vskip -0.1in
\end{table}

\section{$W$ variable x-axis values}

\subsection{Place Values}
 \label{place-list}
Ordered list of bottom 10 and top 10 World Economic Forum Global Gender Gap ranked countries used for the x-axis in \cref{place-multiplot}, that were taken directly without modification from \url{https://www3.weforum.org/docs/WEF_GGGR_2021.pdf}:

    "Afghanistan",
    "Yemen",
    "Iraq",
    "Pakistan",
    "Syria",
    "Democratic Republic of Congo",
    "Iran",
    "Mali",
    "Chad",
    "Saudi Arabia",
    "Switzerland",
    "Ireland",
    "Lithuania",
    "Rwanda",
    "Namibia",
    "Sweden",
    "New Zealand",
    "Norway",
    "Finland",
    "Iceland"

\end{document}